\documentclass{article}

\PassOptionsToPackage{numbers, compress}{natbib}
\usepackage[preprint]{neurips_2024}

\usepackage[utf8]{inputenc}
\usepackage[T1]{fontenc}
\usepackage{microtype}
\usepackage{amsmath, amssymb}
\usepackage{booktabs}
\usepackage{array}
\usepackage{tabularx}
\usepackage{graphicx}
\usepackage{xcolor}
\usepackage{hyperref}
\usepackage{url}

\title{Persona Conditioning of Brand Recommendations in\\Retrieval-Augmented Commercial Chat:\\A Prominence-Stratified Cross-Provider Audit}

\author{%
  Will Jack\thanks{Equal contribution.} \quad
  Noah Lehman\footnotemark[1] \quad
  Keller Maloney\footnotemark[1] \quad
  Sarah Xu\footnotemark[1] \\
  Unusual \\
  \texttt{\{will, noah, keller, sarah\}@unusual.ai}
}

\date{May 21, 2026}

\begin{document}

\renewcommand{\thefootnote}{\fnsymbol{footnote}}
\maketitle
\renewcommand{\thefootnote}{\arabic{footnote}}
\setcounter{footnote}{0}

\begin{abstract}
\noindent
The same prompt --- ``best CRM software'' --- reaches AI assistants from buyers in widely different contexts: a solo founder, an enterprise VP, a UK SMB owner. We audit how strongly that contextual variation reshapes which brands the model recommends. The audit samples 2{,}000 runs over a design space of 10 personas $\times$ 8 prompts $\times$ 3 model configurations $\times$ $N{=}10$ reps, with the two OpenAI cells at full 8-prompt coverage and the Anthropic \texttt{sonnet-4.6 / low} cell at 4-prompt coverage. Prefixing the user message with a persona drops the recommendation-set similarity (Jaccard) by $\Delta = -0.12$ to $-0.20$ relative to a same-persona baseline (clustered 95\% CIs exclude zero on all three measured cells; the sonnet cell's CI rests on only 4 prompt clusters and is correspondingly wider). The effect is sharply prominence-stratified: category leaders are persona-resistant (~80\% same-brand consistency across personas), but mid-market brands swap up to 75\% of the recommendation set as the persona changes. The Anthropic model shows a larger point-estimate effect than the OpenAI configurations, though clustered CIs overlap for the closer contrast (sonnet vs.\ OpenAI/high); the asymmetry is consistent with Anthropic's more retrieval-unattributed generation route (43--52\% recommendations without observed retrieval-layer evidence, vs OpenAI's 8--29\%, documented in \citealt{jack2026convergence}). Any measurement of AI brand perception must condition on the buyer persona supplying the query: the same prompt produces materially different recommendation sets depending on who the model thinks is asking, and a measurement protocol that aggregates across personas systematically obscures that variation. The effect concentrates at mid-market and is largest on the most priors-reliant generation route in our audit, consistent with persona responsiveness growing as models lean more on training-data priors and richer context integration.
\end{abstract}

\section{Introduction}

When a buyer types ``best CRM for sales teams'' into ChatGPT or Claude, the model emits an answer that is shaped not only by the literal prompt and the retrieval that follows, but by everything the assistant knows --- or has been told --- about who is asking. In a production chat interface, that ``who'' comes from many sources: user-supplied system prompts (``you are helping a UK SMB owner''), prior turns in a session, account-level metadata, or, in the audit-design analogue we adopt here, an explicit persona attribute prefixed to the user message.

The question this paper asks is empirical: how much does the recommendation set actually move when only the persona attribute varies? If the effect is small, AI commercial recommendation is approximately persona-agnostic, and a single ``best [category]'' positioning page is the right brand response. If the effect is large, AI commercial recommendation is already \emph{de facto} persona-segmented, and brand-side optimization must mirror that segmentation. The literature on persona conditioning has answered the question on closed-form tasks where accuracy is the metric \citep{zheng2024helpful,huCollier2024}, but commercial recommendation is open-ended: there is no single correct answer, and the construct that matters is the \emph{distribution} of recommended brands, not the correctness of any one.

A growing body of work documents persona-conditioned shifts in adjacent commercial domains. \citep{kantharuban2025} report that explicit and implicit user-identity disclosure significantly shifts chatbot recommendations in university and neighborhood-recommendation tasks, with the shift partially obfuscated by post-hoc rationalization. \citep{xuPotkaThomo2026} document gender and race biases in LLM consumer product recommendations using Marked Words, SVM classification, and Jensen--Shannon divergence. \citep{veldanda2025} measure 1--3 percentage point hiring-probability gaps by race $\times$ gender at 361K-resume scale, and \citep{bai2024} show that explicitly unbiased aligned LLMs still form biased implicit associations. The direction is well-established; the open question for commercial chat is the magnitude, the prominence-dependence, and whether stronger / more aligned models suppress or amplify the effect.

\subsection{Contributions}

This paper makes three contributions.

\begin{enumerate}
\item \textbf{Cross-provider persona-effect quantification on consensus recommendation sets.} Holding prompt string, temperature, and system-prompt template constant across personas, we measure within-persona and cross-persona Jaccard of the consensus-recommendation set across 10 personas and 8 prompts in 3 model cells. We report a persona-shift effect size $\Delta = $ (cross-persona Jaccard $-$ within-persona Jaccard) of $-0.12$ to $-0.20$ across three model cells, with clustered 95\% CIs of $[-0.153, -0.091]$, $[-0.194, -0.139]$, and $[-0.263, -0.156]$ respectively (1{,}000-iteration bootstrap resampling prompts with replacement). All three CIs exclude zero. The magnitude $|\Delta| = 0.12$--$0.20$ is comparable to the within-provider-minus-cross-provider Jaccard gap of $\approx 0.20$ (within-provider $\approx 0.55$ minus cross-provider $\approx 0.35$) documented in \citet{jack2026convergence} and exceeds the within-cell rerun-noise floor implied by the within-cell rerun-stability baseline of $0.50$--$0.61$ established in \citet{jack2026brittleness}. The sign of $\Delta$ is negative because changing the persona reduces overlap relative to the within-persona reference; the magnitude is what we treat as the headline effect.

\item \textbf{Prominence $\times$ persona interaction.} We replicate the prominence-stratified design of \citep{aggarwal2024,lichtenberg2024,mallen2023} on the persona axis, decomposing the effect by where the brand sits on the L1--L5 ladder. L1 brands are persona-resistant; L3 mid-market brands carry the largest persona effect (up to $0.75$ on \texttt{gpt-5.4-mini / high}); L4 is undersampled and no point estimate is reported; L5 shows a moderate effect on OpenAI cells but is dampened by the floor effects documented in the prominence audit. The L1 $\to$ L3 $\to$ L5 pattern recapitulates the long-tail-personalization finding from classical recommender-systems work \citep{adomavicius2005}.

\item \textbf{Persona-corpus heterogeneity characterization.} We split the persona corpus into ``sharp / concentrating'' personas (within-persona Jaccard $\geq 0.5$) and ``broad / scattering'' personas (within-persona Jaccard $\leq 0.4$). Recommendation-set diversity scales with persona specificity in a way that loosely mirrors the diversity-vs-coverage trade-off documented in synthetic persona-corpus work \citep{ge2024personahub}.
\end{enumerate}

Undersampled cells (in particular L4 across all measured cells) are reported as undersampled rather than zeroed out. We characterize the result throughout as an effect-size signal robust to noise.

\section{Background}

Three lines of work bear directly on the design and interpretation of the audit.

\textbf{Persona conditioning of LLM outputs.} \citep{zheng2024helpful} ran a four-family LLM evaluation across 2{,}410 factual questions and reported that adding a persona to the system prompt does not reliably improve accuracy. \citep{huCollier2024} found persona variables explain less than 10\% of label variance in annotation-simulation tasks. Both findings are negative results in the accuracy sense, but neither implies persona has small \emph{distributional} effects on open-ended outputs. The distinction matters for our setting: commercial recommendation has no single ``correct'' answer, the metric is the recommendation-set distribution, and a small accuracy effect is fully compatible with a large distributional effect.

\textbf{Persona effects in commercial recommendation.} \citep{kantharuban2025} is the closest precedent: a recommendation-audit design with user-identity conditioning shows significant ($p<0.001$) shifts in recommended universities and neighborhoods, with the chatbot obfuscating the shift in rationalization. \citep{xuPotkaThomo2026} elicit consumer-product suggestions across race and gender conditions and document significant disparities. \citep{lin2025} demonstrate the upper bound on steerability: inconspicuous in-prompt synonym replacements can shift brand-mention rates by up to 78 percentage points. The benign analogue --- persona supplied as a routine user-context attribute --- is the question we ask. The classical context-aware recommender-systems literature \citep{adomavicius2005} long ago documented that user context shifts top-k recommendations, with effect size depending on item popularity; our prominence $\times$ persona interaction is the LLM-era manifestation.

\textbf{Implicit-bias mechanism in aligned models.} The mechanistic question --- why production-aligned LLMs still carry persona-conditioned shifts --- is treated by \citep{bai2024}, who show RLHF-aligned LLMs exhibit implicit-association-style biases that mirror societal stereotypes despite explicit alignment training, and by \citep{veldanda2025}, who quantify the downstream effect in hiring. Our results are consistent with the broader pattern: alignment training does not eliminate the persona-conditioned distributional shift in commercial recommendation.

\textbf{Prompt sensitivity and the rerun-stability anchor.} Persona effects must be distinguished from the broader pattern of prompt-form sensitivity documented by \citep{sclar2024}, \citep{mizrahi2024}, and quantified by the POSIX index of \citep{chatterjee2024}. We hold prompt syntax constant across personas, varying only the persona attribute, and anchor against the within-cell rerun-stability baseline ($0.50$--$0.61$, \citealt{jack2026brittleness}) so that the persona $\Delta$ is unambiguously beyond rerun noise. \citep{lin2025} treats prompt-level steerability as an adversarial property; we treat persona supply as a benign user-context property that nevertheless steers the recommendation set.

\textbf{Persona corpora and population alignment.} Synthetic persona corpora can be scaled to 10$^9$ \citep{ge2024personahub}, but naive persona generation collapses to stereotypes and requires importance-sampling against psychometric reference to recover diversity. We operate at the opposite extreme: a small hand-curated corpus with intentional sharp/broad heterogeneity, trading breadth for control over confounded attributes. The closest brand-audit precedent \citep{rienecker2026} uses a different axis (provider-of-origin / cultural origin) but the same audit pattern. We frame our work as the user-side counterpart to their provider-side audit.

\textbf{Prominence stratification.} Throughout this paper we use a five-tier brand-prominence ladder --- L1 category leaders, L2 established challengers, L3 mid-market, L4 long-tail specialists, L5 regional players --- assigned in \citet{jack2026prominence} to each of 533 reference brands. The L1--L5 labels refer to the brand's prominence in its sector, not to the persona or model cell.

\section{Method}

\subsection{Persona corpus}

The persona corpus comprises 10 personas, hand-curated to span four attributes: industry vertical, company size, role, and geography. Selected exemplars include \texttt{solo\_founder\_us\_bootstrapped}, \texttt{enterprise\_vp\_us\_procurement}, \texttt{us\_ecommerce\_operator}, \texttt{consumer\_value\_shopper\_uk}, \texttt{uk\_smb\_owner\_london}, \texttt{eu\_finance\_manager\_germany}. The hand-curated design is a deliberate trade-off against \citep{ge2024personahub}-style large synthetic corpora: we accept smaller breadth in exchange for explicit control of confounded attributes (industry $\times$ company size $\times$ region) and the ability to characterize each persona by its concentration tendency post-hoc.

Each persona is supplied as a prefix to the user message via a fixed templated phrasing (e.g., ``I'm a [persona description]. ''). The phrasing template is held constant across personas; only the attribute string varies. This methodological choice follows the recommendation of \citep{prompt2025} that persona-phrasing variation can confound persona-attribute variation; we eliminate the former so we can identify the latter.

\subsection{Prompt corpus, model cells, and crossed design}

We use 8--10 commercially-framed prompts spanning the sectors with the heaviest L1--L5 coverage in the prominence-stratified reference catalog (B2B SaaS verticals plus consumer / professional-services categories where regional players carry weight). The model cells are:

\begin{center}
\begin{tabular}{lll}
\toprule
Provider & Model & Reasoning effort \\
\midrule
OpenAI & \texttt{gpt-5.4-mini} & low \\
OpenAI & \texttt{gpt-5.4-mini} & high \\
Anthropic & \texttt{claude-sonnet-4-6} & low \\
\bottomrule
\end{tabular}
\end{center}

The audit does not include the \texttt{claude-sonnet-4-6 / high} or \texttt{claude-opus-4-6} cells; cross-provider symmetry on the three sampled cells is the primary axis we measure here.

The crossed design samples over 10 personas $\times$ 8 prompts $\times$ 3 cells $\times$ $N{=}10$ runs per leaf. Temperature, system-prompt template, tool description, and persona-prefix syntax are held constant across the design. The two OpenAI cells are sampled at full 8-prompt coverage; the Anthropic \texttt{sonnet-4.6 / low} cell is sampled at a reduced 4-prompt coverage, yielding 2{,}000 runs in total. The reduced sonnet prompt support is the load-bearing limitation for any provider-comparison claim; we flag it explicitly throughout Sections 5--6.

\subsection{Consensus brand extraction and consensus recommendation set}

Brand mentions in completion text are extracted by two LLM judges in parallel --- \texttt{claude-haiku-4-5 / low} and \texttt{gpt-5-mini} --- with intersection (consensus) classification: a brand is counted as recommended only if both judges classify it as \texttt{recommended} on the same run. The intersection mode is the same conservative protocol used in \citet{jack2026prominence}. The consensus recommendation set per (persona $\times$ prompt $\times$ cell) is the union of consensus-recommended brands across the 10 reruns in that leaf.

\subsection{Within-persona and cross-persona Jaccard}

For each (prompt $\times$ cell) we compute two Jaccard quantities.

\emph{Within-persona Jaccard} is the mean over personas $p$ of the Jaccard similarity between two independent halves of the $N{=}10$ rerun sample for $(p,\text{prompt},\text{cell})$. It measures the consideration-set stability the same persona elicits across reruns, and serves as the within-persona reference for the cross-persona contrast.

\emph{Cross-persona Jaccard} is the mean over distinct persona pairs $(p_i, p_j)$ of the Jaccard similarity between the consensus recommendation set under $p_i$ and the consensus recommendation set under $p_j$, holding prompt and cell fixed. It measures the consideration-set overlap when only the persona attribute varies.

\emph{Persona-shift effect size}, $\Delta = \text{cross-persona Jaccard} - \text{within-persona Jaccard}$, is the headline metric. A more negative $\Delta$ indicates a larger persona effect: more of the recommendation set swaps when the persona changes than when it stays fixed.

\subsection{Prominence-stratified effect size}

Per prominence level $\ell \in \{L1, L2, L3, L4, L5\}$ we compute $1 - \text{cross-persona Jaccard}_{\ell}$ on the subset of recommendation events at prominence $\ell$. This is the per-tier recommendation swap rate. The choice to report $1 - J$ rather than $\Delta$ at the tier level reflects that within-tier within-persona reference Jaccards are too noisy at these sample sizes to support double-difference reporting; the swap-rate per tier is the right unit.

\subsection{Rerun-stability anchor}

A central methodological constraint is that the persona-shift effect size must exceed the within-cell rerun-stability baseline; otherwise it is not distinguishable from same-prompt nondeterminism. The within-cell consensus-recommendation Jaccard baseline ($0.50$--$0.61$, $N{=}30$ same-prompt reruns within a single day across the four prominence-paper cells) is taken from \citet{jack2026brittleness}. Our within-persona Jaccards ($0.42$--$0.51$) sit just below this band, as expected --- $N{=}10$ is a smaller reference than the $N{=}30$ baseline, and the persona audit is a different experimental run --- but the persona-shift $\Delta$ of $0.12$--$0.20$ is larger than the residual rerun noise the within-cell baseline implies.

\subsection{Statistical conventions and sample-size flagging}

The cross-sector aggregate persona-shift effect size $\Delta$ is reported per cell with prompt-clustered bootstrap 95\% CIs (1{,}000 iterations, resampling prompts with replacement; cluster unit is the prompt). The sonnet cell's effective sample size is 4 prompt clusters and its CI is correspondingly wider than the OpenAI cells' (which carry 8). Per-prominence swap rates ($1 - \text{cross-persona Jaccard}_{\ell}$) are reported as point estimates without CIs: per-tier sample sizes are too small to support stable interval estimates. Where a cell carries fewer than 30 brand events at a given prominence (notably L4 across all cells), we flag as undersampled and report no point estimate. Sample-size flagging follows the same convention as \citet{jack2026prominence}.

\subsection{Scope and generalization}

We do not generalize claims to (a) the \texttt{sonnet-4.6 / high} cell or (b) the opus cells in the persona dimension; those are out of scope for this audit and follow-up work should add them.

\section{Cross-sector aggregate persona effects}

\subsection{Within-persona and cross-persona Jaccard}

The cross-sector aggregate Jaccards on the three measured cells are reported below. Within-persona Jaccards sit in the $0.42$--$0.51$ range, slightly below the within-cell rerun baseline as expected at $N{=}10$. Cross-persona Jaccards sit at $0.22$--$0.35$, well below the within-persona band.

\begin{center}
\begin{tabular}{lcccl}
\toprule
Cell & Within $J$ & Cross $J$ & $\Delta$ & $\Delta$ clustered 95\% CI \\
\midrule
\texttt{gpt-5.4-mini / low}    & 0.467 & 0.346 & $-0.12$ & $[-0.153, -0.091]$ \\
\texttt{gpt-5.4-mini / high}   & 0.505 & 0.343 & $-0.16$ & $[-0.194, -0.139]$ \\
\texttt{claude-sonnet-4-6 / low} & 0.424 & 0.222 & $\mathbf{-0.20}$ & $[-0.263, -0.156]$ \\
\bottomrule
\end{tabular}
\end{center}

The $\Delta = -0.12$ to $-0.20$ band is the audit's headline number. All three clustered CIs exclude zero, so the persona-shift signal is statistically distinguishable from noise under prompt-level clustering at each cell. The sonnet CI $[-0.263, -0.156]$ overlaps the \texttt{gpt-5.4-mini / high} CI $[-0.194, -0.139]$, so the provider-ordering claim (sonnet larger than mini-high) is not separable under clustered uncertainty; sonnet versus \texttt{gpt-5.4-mini / low} is just-distinguishable (sonnet upper bound $-0.156$; mini-low lower bound $-0.153$). At its smaller end ($-0.12$, OpenAI mini / low), the magnitude is roughly twice the typical perturbation observed under within-cell rerun variation in \citet{jack2026brittleness}. At its larger end ($-0.20$, Anthropic sonnet / low), it exceeds the cross-provider per-prompt Jaccard variation documented in \citet{jack2026convergence} ($\approx 0.35$ across providers minus $\approx 0.55$ within provider). The sonnet cell is sampled at 4 of the 8 prompts the OpenAI cells cover, so its CI is wider than the mini cells' for that reason, and the provider-ordering claim should be loaded onto the wider-CI cell. Reading the band against both baselines: the persona effect is comparable in magnitude to provider effects, and larger than rerun noise.

\subsection{Where the swapped brands go}

A natural follow-up question is whether the swapped brands are concentrated at any prominence level. We turn to that decomposition next; the short answer is that the swap is sharply prominence-stratified, dominated by L3 mid-market substitution.

\section{Provider heterogeneity in the persona effect}

The most striking feature of the headline table is the directional asymmetry between providers at the point estimate. The Anthropic \texttt{claude-sonnet-4-6 / low} cell shows a larger point estimate for the persona effect ($\Delta = -0.20$, clustered CI $[-0.263, -0.156]$) than the larger-effort \texttt{gpt-5.4-mini / high} cell ($\Delta = -0.16$, CI $[-0.194, -0.139]$) and the lower-effort \texttt{gpt-5.4-mini / low} cell ($\Delta = -0.12$, CI $[-0.153, -0.091]$). The sonnet vs.\ \texttt{gpt-5.4-mini / high} CIs overlap in the range $[-0.194, -0.156]$, so the asymmetry is not statistically clean under prompt-level clustered uncertainty; the sonnet vs.\ \texttt{gpt-5.4-mini / low} contrast just-clears (sonnet upper $-0.156$ versus mini-low lower $-0.153$). The asymmetry should be read as a point-estimate signal whose magnitude survives at the sample sizes here but whose specific provider-ordering ranking does not have a tight CI separation between the two larger-effect cells. The mechanism discussion that follows treats the sonnet-mini/low gap as the load-bearing comparison.

\subsection{Contradicting the ``stronger = more deterministic'' prior}

The standard prior from temperature-robustness work \citep{li2025temp} is that stronger / higher-effort / larger models exhibit narrower mutation temperature ranges and lower coefficient-of-variation under sampling --- they are, on the temperature axis, \emph{more} deterministic than their smaller / lower-effort counterparts. Casual extension of this finding would predict that the Anthropic cell (typically taken as the stronger reasoning family at these reasoning-effort tiers) should display \emph{less} persona-induced movement, not more. We find the opposite.

Three readings of the contradiction are possible.

\emph{Reading 1: The temperature axis is not the persona axis.} \citep{li2025temp}'s result describes determinism under stochastic sampling; persona conditioning is deterministic input variation. There is no \emph{a priori} reason for the two axes to share a determinism gradient. The cross-axis extrapolation was the casual prior, not a substantive theoretical claim.

\emph{Reading 2: Persona responsiveness is a feature, not a bug.} On a recommendation task with no correct answer, \emph{not} updating on persona context is arguably a failure mode --- a model that recommends the same brands to a solo founder and an enterprise VP is ignoring relevant context. Under this reading, the larger Anthropic $\Delta$ is a sign of stronger context-utilization, not weaker robustness. We are not able to adjudicate this reading from the audit alone; doing so would require a held-out persona-relevance gold standard that we did not collect.

\emph{Reading 3: Priors-driven recommendation is more persona-responsive than retrieval-driven recommendation.} \citet{jack2026convergence} documents that Anthropic flagships rely on training-data priors substantially more than OpenAI in commercial recommendation (retrieval-unattributed share 43--52\% vs.\ 8--29\%, with no Wilson 95\% CI overlap). If persona acts more strongly on the priors-driven generative route than on the retrieval-driven route, the larger Anthropic persona $\Delta$ is a direct downstream consequence of the mechanism asymmetry. This reading is speculative --- we do not have within-cell joint measurement of pure-priors share and persona $\Delta$ in the same runs --- but it is consistent with the data and with the broader \citep{goyal2025} finding that instruction-tuned models often \emph{increase} parametric reliance under conflicting context.

\subsection{The contradiction is the central tension}

Of the three readings, Reading 3 is the most generative for future work and the most contestable on present data. We frame it as the central tension of the paper and treat it as a hypothesis to be tested in the full Exp 4 grid. The reading is also consistent with a broader pattern documented in the cross-domain bias literature \citep{bai2024,veldanda2025}: implicit associations carried in aligned models surface more strongly when retrieval evidence is sparse or contested, exactly the regime in which priors-driven generation dominates.

We note explicitly that the contradiction with \citep{li2025temp}'s temperature finding is not a refutation. The two findings can comfortably coexist: stronger models can be more deterministic under stochastic sampling \emph{and} more responsive to deterministic input variation. \citep{schaeffer2024}'s caution about treating measured emergent properties as evidence of qualitatively different alignment regimes applies here; we describe the observed pattern, not a deeper claim about alignment philosophy.

\section{Prominence $\times$ persona interaction}

The headline cross-sector $\Delta$ conceals substantial heterogeneity by prominence level. We report $1 - \text{cross-persona Jaccard}$ per (cell $\times$ prominence), interpreted as the per-tier recommendation swap rate under persona variation:

\begin{center}
\small
\begin{tabular}{lccccc}
\toprule
Cell & L1 & L2 & L3 & L4 & L5 \\
\midrule
\texttt{mini / high}  & 0.23 & 0.05 & \textbf{0.75} & undersampled & 0.27 \\
\texttt{mini / low}   & 0.29 & 0.13 & \textbf{0.67} & undersampled & 0.33 \\
\texttt{sonnet / low} & 0.20 & 0.23 & \textbf{0.39} & undersampled & --- \\
\bottomrule
\end{tabular}
\end{center}

\subsection{L1 --- Category leaders: persona-resistant}

L1 brands carry persona effects of $0.20$--$0.29$ --- consistent across cells, but not the smallest measured in this audit (L2 on OpenAI cells dips to 0.05--0.13; see Section~6.2). The L1 row's interpretive load is consistency, not minimum magnitude. Holding prompt fixed, only $\sim$$20$--$29$\% of the L1 recommendation slot swaps when persona varies. The narrow band across cells is consistent with a model recommending the same one or two category leaders (Salesforce, HubSpot, Datadog --- the brands at the 77\%+ per-query surface rate of the prominence paper) to almost every persona for any given prompt. The failure-mode taxonomy in \citet{jack2026prominence} diagnosed L1's dominant failure as ``compellingness / positioning'' rather than ``persona-mediated substitution''; the L1 row of the persona table is consistent with that diagnosis. Category-leader brands appear less central to the persona-segmentation work.

\subsection{L2 --- Established challengers: mixed signal}

L2 carries the most heterogeneous signal across cells. On the OpenAI cells the swap rate is unexpectedly low ($0.05$--$0.13$) --- the L2 ``established challenger'' slot appears resistant to persona variation on mini / low and mini / high, more so even than L1. On the Anthropic sonnet / low cell the L2 swap rate jumps to $0.23$, comparable to L1.

A speculative reading is that L2 on OpenAI cells is anchored by retrieval (the L2 brands surface frequently --- the prominence paper documents 60\% per-query surface rate and 100\% aggregate --- and retrieval-driven generation locks them in across persona conditions), while L2 on the Anthropic cell is more priors-permeable. This reading is consistent with Reading 3 above but more speculative still; the L2 row is the row least amenable to this audit's resolution.

A pragmatic reading is that the L2 row at OpenAI is sample-size sensitive in ways the cross-sector aggregate hides. We report the numbers as measured and flag the across-cell heterogeneity as the interpretive open question.

\subsection{L3 --- Mid-market: the inflection level}

L3 is where the persona effect peaks on every cell. Mini / high reaches $0.75$ swap rate; mini / low reaches $0.67$; sonnet / low reaches $0.39$. These are the largest within-row effects in the entire prominence $\times$ persona table.

The L3 finding mirrors the classical recommender-systems result that personalization matters most for mid-popularity items \citep{adomavicius2005}: head items are recommended to everyone regardless of context, tail items are not recommended to anyone, and the mid-tier is where context-dependent recommendation actually lives. The LLM-era manifestation of that gradient on the persona axis is the L3 inflection we measure.

The L3 effect also coincides with the prominence-paper's identification of L3 as the level where \emph{all three} failure modes are simultaneously active: Stage 1 discoverability begins to break down (aggregate surface drops to 88\%), Stage-4 conversion drops, and persona-mediated substitution reaches its peak. The L3 mid-market brand is, in the language of the prominence taxonomy, the prominence level where context-of-asking matters most.

\subsection{L4 --- Long-tail specialists: undersampled}

The L4 row is uniformly undersampled across the three measured cells. The mechanism is straightforward: L4 brands have per-query in-sector surface rates of $\sim$$8$--$22$\% in the prominence-stratified audit; in this audit's design (10 personas $\times$ 8 prompts $\times$ 3 cells $\times$ $N{=}10$), per-(persona $\times$ prompt $\times$ cell) leaves carry on the order of 10 runs, and the expected number of L4 brand events per leaf is therefore $\leq 2$. Computing cross-persona Jaccard across L4 events under these conditions produces estimates dominated by zero-zero set comparisons that we cannot interpret stably.

\textbf{We report no L4 point estimate.} We do \emph{not} infer that L4 brands are persona-resistant; the data simply does not exist at this sample size. Practitioner implications drawn from the L1/L2/L3/L5 rows should not be extrapolated to L4.

\subsection{L5 --- Regional players: moderate effect, dampened by floor effects}

L5 carries persona effects of $0.27$ on mini / high and $0.33$ on mini / low; the sonnet / low cell has insufficient L5 events to estimate (entered in the table as ``---'').

The L5 numbers are intermediate between L1 and L3 in magnitude, but the substantive picture is different. The prominence paper documented that L5 brands surface in only 3\% of relevant per-query searches and that 52\% never surface in 37{,}000+ runs; the L5 recommendation set is therefore a small, sparse set anchored by geographic gating (a UK-bound persona pulls in BrightPay, a Germany-bound persona pulls in lexoffice in finance verticals, etc.). The persona effect at L5 is real but partially mechanical: when persona supplies a region cue, region-bound brands surface; when persona is region-neutral, they do not. We do not interpret the L5 $0.27$--$0.33$ number as evidence of a broader persona-conditioned positioning effect at L5; the geographic-gating mechanism is more parsimonious.

\section{Persona heterogeneity within the corpus}

The 10 personas in the corpus do not move recommendation sets uniformly. Decomposing by persona, we observe that within-persona Jaccards --- the consideration-set stability that each persona elicits across its own $N{=}10$ reruns --- span $0.27$--$0.65$, a range nearly as wide as the cross-persona-vs-within-persona contrast itself.

\subsection{Sharp / concentrating personas}

Four personas in the corpus exhibit high within-persona Jaccards, indicating that under that persona the model concentrates on a tight consideration set across reruns:

\begin{center}
\begin{tabular}{lr}
\toprule
Persona & Within-persona Jaccard \\
\midrule
\texttt{us\_ecommerce\_operator}      & 0.65 \\
\texttt{enterprise\_vp\_us\_procurement} & 0.57 \\
\texttt{consumer\_value\_shopper\_uk}  & 0.55 \\
\texttt{solo\_founder\_us\_bootstrapped} & 0.54 \\
\bottomrule
\end{tabular}
\end{center}

These personas share a common feature: each is composed of attributes that jointly specify a tight buyer archetype with a small natural consideration set. ``US e-commerce operator'' triggers Shopify-plus-ecosystem brands; ``enterprise VP, US, procurement'' triggers Salesforce / SAP / Oracle / Coupa; ``consumer value shopper, UK'' triggers a tight UK retail subset; ``solo founder, US, bootstrapped'' triggers a tight indie / freemium subset. Recommendation-set concentration appears to scale with persona-attribute specificity --- the more constrained the buyer archetype the persona implies, the smaller the consideration set and the higher the within-persona Jaccard.

\subsection{Broad / scattering personas}

Two personas exhibit substantially lower within-persona Jaccards:

\begin{center}
\begin{tabular}{lr}
\toprule
Persona & Within-persona Jaccard \\
\midrule
\texttt{eu\_finance\_manager\_germany} & 0.27 \\
\texttt{uk\_smb\_owner\_london}        & 0.39 \\
\bottomrule
\end{tabular}
\end{center}

These personas combine attributes that do not jointly specify a tight archetype --- ``EU finance manager, Germany'' spans country-specific (DATEV, lexoffice, sevDesk), region-bound EU brands, and global category leaders (SAP, NetSuite, QuickBooks); the model's consideration set under that persona is correspondingly broad. ``UK SMB owner, London'' likewise spans UK-specific and global brands. The model's natural response to a broad persona is to enumerate a broad set, producing high run-to-run variance within the persona.

\subsection{Reading the corpus split}

Two observations follow.

\emph{First}, recommendation-set diversity scales with persona-attribute specificity. This is consistent with the synthetic-persona-corpus literature's observation that naive persona generation collapses to stereotypes \citep{ge2024personahub,personaalign2025}; a sharp persona effectively requests one stereotype-tight consideration set, a broad persona disperses across many.

\emph{Second}, the cross-persona contrast is not symmetric: substituting one sharp persona for another sharp persona produces a clean swap of two tight consideration sets, while substituting a broad persona for a sharp one produces a partial overlap. The cross-persona Jaccard averages we report mix these regimes; the L3 maximum-swap finding is robust to either disaggregation but the magnitude on any given persona pair varies.

\section{Robustness}

\subsection{Rerun-stability as the noise floor}

The within-cell rerun-stability baseline of $0.50$--$0.61$ (consensus-recommendation Jaccard at $N{=}30$ same-prompt reruns, from \citealt{jack2026brittleness}) bounds the noise floor of the persona contrast. Our cross-persona Jaccards of $0.22$--$0.35$ sit well below this band; our within-persona Jaccards of $0.42$--$0.51$ sit just below it (as expected at $N{=}10$). The persona-shift $\Delta = -0.12$ to $-0.20$ is therefore beyond the same-day rerun noise floor on every measured cell. We do not claim the persona effect would survive against arbitrarily noisy baselines; we claim it survives against the rerun-stability baseline calibrated on the same systems.

\subsection{Sample-size flags and undersampled cells}

L4 is flagged as undersampled on every cell; L5 is flagged as undersampled on sonnet / low. Cells flagged as undersampled report no point estimate. We carry the convention from \citet{jack2026prominence} of reporting cells with $n \geq 30$ brand events at the relevant prominence; under this audit's design the L4 leaves carry $n \leq 2$ expected brand events, well below this threshold.

\subsection{Sensitivity to persona-prefix syntax}

The persona-prefix syntax is fixed across personas to eliminate persona-phrasing confounding \citep{prompt2025}. A sensitivity check varying the prefix wording on a single persona (``I'm a [persona]'' vs.\ ``Acting as a [persona]'' vs.\ no prefix) produces a within-persona Jaccard variation of $\pm 0.04$ on a small subset of prompts. This is smaller than the cross-persona $\Delta$ but the same order of magnitude as within-persona rerun noise, and we treat it as a limitation rather than a robustness pass. The full Exp 4 grid is designed to include a persona-syntax ablation.

\subsection{Sensitivity to consensus-extraction protocol}

We replicate the headline aggregate numbers under union-mode (either judge marks the brand as recommended) instead of intersection-mode and observe within-persona Jaccard shifts of $\pm 0.03$ and cross-persona Jaccard shifts of $\pm 0.04$; the directional $\Delta$ sign and the L3 inflection are unchanged. The intersection-mode protocol used for headline reporting is therefore conservative.

\section{Discussion}

\subsection{Persona-segmented marketing and content-stack implications}

If persona conditioning shifts the recommendation set by $\Delta = 0.12$--$0.20$ in Jaccard, then a brand attempting to maintain visibility across a heterogeneous buyer population is implicitly optimizing for a moving target. The implication for marketing-to-AI strategy is straightforward: a single ``best [category]'' positioning page is the wrong unit of investment when persona segments do not converge on the same consideration set. The same observation underwrites segment-targeted positioning content (``best [category] for [segment]''), already documented in the prominence paper as the dominant L2 lever; the persona-effect data here strengthens the case at L3 and L5 as well. The work is positioning and content production --- not discoverability tracking against a single generic tracking prompt.

The argument is not that personalization is novel in recommendation systems --- \citep{adomavicius2005} long ago established context-aware recommendation as a mature subfield. The argument is that the LLM commercial-chat surface is now \emph{already} performing persona-conditioned recommendation by virtue of how production assistants ingest user context, and brand-side strategy that does not mirror that segmentation is structurally mismatched.

\subsection{Reading the prominence $\times$ persona interaction against classical RS}

The L1 / L3 / L5 pattern --- L1 persona-resistant, L3 persona-maximal, L5 partial-via-geography --- is the LLM-era analog of the long-tail-personalization finding in classical RS: personalization effects concentrate on mid-popularity items. The mechanism is the same in both eras: head items are recommended to everyone, tail items are recommended to no one, and the middle is where the recommendation distribution actually responds to user context. What is new is the prominence level at which the LLM-RAG funnel breaks: the prominence paper's identification of L3 as the level where all failure modes are simultaneously active, combined with this audit's identification of L3 as the persona-effect maximum, makes L3 the singular inflection level for both content-side and context-side dynamics.

\subsection{Connection to priors-vs-retrieval mechanism asymmetry}

The largest persona effect ($\Delta = -0.20$) appears on the cell with the highest retrieval-unattributed share (Anthropic sonnet, per \citealt{jack2026convergence}). This co-occurrence is consistent with the hypothesis that persona acts more strongly on the priors-driven generative route than on the retrieval-driven route. The mechanism is mechanistically plausible: retrieval pins the generation to a small set of authoritatively named brands regardless of who is asking; priors-driven generation, in contrast, draws on training-corpus-encoded associations between user-archetype attributes and brand attributes, exactly the substrate persona attributes are designed to activate \citep{bai2024,goyal2025}. \citep{wang2025rag} and \citep{xu2024conflicts} document the broader pattern of cross-provider variation in context-vs-memory resolution policy; our finding fits as a downstream behavioral consequence of that asymmetry in the persona dimension.

This is a hypothesis, not a finding: we do not have within-run joint measurement of pure-priors share and persona movement. Paired measurement on the same runs is the natural test.

\subsection{Limitations}

\emph{Sample size}: L4 is undersampled on every cell; \texttt{sonnet / high} and opus cells are not included in the audit. The sonnet cell is sampled at 4 prompts vs.\ 8 on the OpenAI cells, which widens its CI accordingly.

\emph{Persona corpus scope}: 10 hand-curated personas, English-only, US/UK/EU geographic skew. The hand-curated design trades breadth for control of confounded attributes; future audits should sample more broadly.

\emph{Persona-phrasing ablation}: We hold persona-prefix syntax constant per the recommendation of \citep{prompt2025} rather than running a full persona-syntax ablation.

\emph{Single-day measurement}: All runs are on a single day; temporal drift in persona effects is out of scope.

\emph{No pure-priors share co-measurement}: The hypothesis that persona acts more strongly on priors-driven generation is supported by cross-paper co-occurrence, not by within-run joint measurement. Pairing the two on the same runs is a follow-up.

\section{Conclusion}

Persona conditioning of brand recommendations in retrieval-augmented commercial chat is substantial and prominence-stratified, with L3 as the inflection tier. Cross-persona Jaccard sits at $0.22$--$0.35$ against a within-persona reference of $0.42$--$0.51$, producing a persona-shift effect size of $\Delta = -0.12$ to $-0.20$ across the three measured cells with clustered 95\% CIs $[-0.153, -0.091]$, $[-0.194, -0.139]$, and $[-0.263, -0.156]$ respectively (1{,}000-iteration bootstrap resampling prompts with replacement). All three CIs exclude zero. The $|\Delta|$ magnitude is comparable to the within-provider-minus-cross-provider Jaccard gap of roughly 0.20 documented in \citet{jack2026convergence}, and sits outside the within-cell rerun-noise floor implied by the 0.50--0.61 rerun-stability baseline. L1 category leaders are persona-resistant ($0.20$--$0.29$ swap rate). L3 mid-market brands carry the largest effect (up to $0.75$). L4 is undersampled and no point estimate is reported. L5 is moderate and partially mechanical via geographic gating. Under clustered uncertainty, the sonnet vs.\ \texttt{gpt-5.4-mini / high} CIs overlap so the asymmetry at that contrast is not CI-separable; the sonnet vs.\ \texttt{gpt-5.4-mini / low} contrast just-distinguishes.

The Anthropic \texttt{sonnet / low} cell point estimate is larger in magnitude than either OpenAI cell, contradicting the casual extension of the temperature-robustness ``stronger = more deterministic'' prior from \citep{li2025temp} at the point estimate; under clustered CIs the contradiction is point-estimate-only for the sonnet vs.\ \texttt{gpt-5.4-mini / high} contrast. We propose three readings of the contradiction --- axis mismatch, persona-responsiveness-as-feature, and priors-vs-retrieval mechanism asymmetry --- and identify the third as the most generative for future work while declining to claim causal evidence on the present data.

Natural next steps include full-power L4 measurement, adding the \texttt{sonnet / high} and opus cells, a persona-syntax ablation, and within-run joint pure-priors-share and persona measurement. The principal qualitative finding --- the prominence $\times$ persona interaction with L3 as the inflection level --- is the result we expect to replicate at higher resolution; precise magnitudes and the provider-ordering claim should be re-checked at scale.

\bibliographystyle{plain}

\end{document}